\documentclass[runningheads]{llncs}

\usepackage{eccv}

\usepackage{eccvabbrv}

\usepackage{graphicx}
\usepackage{booktabs}

\usepackage[accsupp]{axessibility}  

\usepackage{hyperref}
\hypersetup{
	hidelinks,
	colorlinks=true,
}

\usepackage{orcidlink}

\usepackage{comment}
\usepackage{amsmath,amssymb} 
\usepackage{color}
\usepackage{multirow}
\usepackage{ulem}

\begin{document}

\title{PISR: Polarimetric\protect\\
	Neural Implicit Surface Reconstruction\protect\\
	for Textureless and Specular Objects} 

\titlerunning{PISR for Textureless and Specular Objects}

\author{Guangcheng Chen\inst{1,2}\orcidlink{0000-0001-8191-3327} \and
Yicheng He\inst{1,2}\orcidlink{0000-0003-3213-9907} \and
Li He\inst{1,2}\orcidlink{0000-0003-0261-4068} \and
Hong Zhang\thanks{Corresponding author.}\inst{1,2}\orcidlink{0000-0002-1677-6132}}

\authorrunning{G. Chen et al.}

\institute{Southern University of Science and Technology, Shenzhen, China\and
Shenzhen Key Laboratory of Robotics and Computer Vision
\email{\{chengc2023,12332136\}@mail.sustech.edu.cn}, \email{\{hel, hzhang\}@sustech.edu.cn}}

\maketitle

\begin{abstract}
Neural implicit surface reconstruction has achieved remarkable progress recently. Despite resorting to complex radiance modeling, state-of-the-art methods still struggle with textureless and specular surfaces. Different from RGB images, polarization images can provide direct constraints on the azimuth angles of the surface normals. In this paper, we present PISR, a novel method that utilizes a geometrically accurate polarimetric loss to refine shape independently of appearance. In addition, PISR smooths surface normals in image space to eliminate severe shape distortions and leverages the hash-grid-based neural signed distance function to accelerate the reconstruction. Experimental results demonstrate that PISR achieves higher accuracy and robustness, with an L1 Chamfer distance of $0.5$ mm and an F-score of $99.5$\% at $1$ mm, while converging $4\sim 30\times$ faster than previous polarimetric surface reconstruction methods.
\let\thefootnote\relax\footnotetext{The source code is available at \url{https://github.com/GCChen97/PISR}}

\end{abstract}

\section{Introduction}
\label{sec:intro}

Surface reconstruction from multi-view images is a fundamental problem in computer vision,
with various applications in graphics, robotics, and more.
Recently, neural implicit surface has shown powerful flexibility for representing 3D shape and appearance.
Combined with differentiable volume rendering \cite{mildenhall2021nerf},
neural implicit surface reconstruction is able to estimate 3D shapes from multi-view images accurately \cite{wang2021neus,yariv2021volume}.
However, these methods struggle with textureless and specular objects due to the shape-radiance ambiguity \cite{kaizhang2020}.
Recent methods resort to diffuse and specular separation modeling \cite{wang2023unisdf},
specular reflection detection \cite{ge2023ref} and inverse rendering techniques \cite{physg2021, liu2023nero}.
While these methods show promising results, they may still encounter difficulties
since their optimizations rely on image reconstruction losses that tightly couple shape and appearance.

Polarization, as another property of light, provides direct cues on object shape.
The angle of polarization of light from object surfaces is independent of light intensity and mainly determined by the surface shapes.
From the perspective of a camera, the angle of polarization of a pixel is equal to the azimuth of the surface normal, i.e., the 2D projection direction of the normal onto the image plane, up to a $\pi/2$-ambiguity and a $\pi$-ambiguity \cite{cui2017polarimetric}.
This geometric property has been extensively explored in
shape from polarization \cite{miyazaki2003polarization, atkinson2006recovery, yu2017shape, ba2020deep}, 
photo-polarimetric stereo \cite{ngo2015shape, tozza2021uncalibrated, ding2021polarimetric, ding2022polar}, polarimetric multi-view 3D reconstruction \cite{cui2017polarimetric, zhao2022polarimetric,zhao2024polarimetric} and dense mapping \cite{yang2018polarimetric, shakeri2021polarimetric}.

In this work, we introduce PISR (polarimetric neural implicit surface reconstruction), a novel method that combines the merits of polarimetric 3D reconstruction and neural implicit surface reconstruction for textureless and specular objects, such as ceramics and plastics.
Different from recent works \cite{dave2022pandora, li2023neisf} that utilize the polarization cues through polarimetric volume rendering,
we propose a novel polarimetric loss to refine an object shape directly.
\textcolor{black}{
Thanks to the integration of the perspective polarimetric constraint \cite{chen2022perspective} in the proposed loss, PISR eliminates artifacts on the reconstructed surfaces, resulting in a 30\% reduction in L1 Chamfer distance.
}Besides, instead of treating all possible normals equally \cite{zhao2022polarimetric},
\textcolor{black}{
our loss adaptively adjusts according to the degree of polarization to address specular-reflection-dominant surfaces, such as ceramics with dark color.
}

\begin{figure}[t]
	\centering
	\includegraphics[width=122mm]{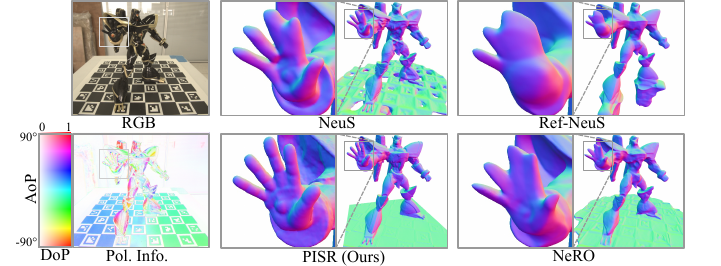}
	\caption{
		Normal maps of reconstructed surfaces.
		By leveraging polarization information, our method is able to reconstruct textureless and specular surfaces with fine-grained details.
	}
	\label{fig:ch1-intro}
\end{figure}

To accelerate the optimization process, PISR uses a multi-resolution hash grid and multi-layer perceptrons (MLP) to represent an object, achieving $4\sim30\times$ speed up as shown in \cref{table:speed}.
This representation is also crucial for accurately reconstructing highly reflective surfaces, enabling PISR to rectify the shape topology during optimization,
a processing step not available for the mesh-based method, PMVIR \cite{zhao2022polarimetric}.
However, the discrete nature of a hash grid also reduces the smoothness of the representation and can lead to the reconstruction results being trapped in local minima.
To mitigate the problem, PISR smooths rendered normals for better robustness against poor shape initialization.

\textcolor{black}{
Since there is no publicly available dataset for quantitative evaluation of polarimetric 3D reconstruction,
}we collect a dataset containing textureless and specular objects with a color polarization camera \cite{polsensor}.
Using the dataset, we show that
PISR is able to improve the surface reconstruction accuracy over the SOTA methods, with an L1 Chamfer distance of $0.5$ mm and an F-score of $99.5$\% at $1$ mm.
Shown in \cref{fig:ch1-intro} are recovered shapes by different methods and our result shows an accurate shape and fine-grained details.

In summary, our contributions are as follows:
\begin{itemize}
	\item
	PISR, a novel method that leverages polarization cues and hash-grid-based neural SDF for robust and efficient reconstruction of textureless and specular objects.
	\item
	A novel polarimetric loss that utilizes an accurate formulation of the polarimetric constraint for higher surface reconstruction accuracy.
	\item
	A polarization image dataset of real textureless and specular objects with ground-truth meshes for quantitative evaluation.
\end{itemize}

\section{Related Works}

\subsubsection{Neural Implicit Surface Reconstruction:}
The success of neural radiance field reconstruction \cite{mildenhall2021nerf} in novel view synthesis greatly inspires the development of neural implicit surface \cite{park2019deepsdf, yariv2020multiview, DVR2020cvpr}.
Combining with differentiable volume rendering, neural implicit surface reconstruction \cite{wang2021neus, yariv2021volume, Oechsle2021ICCV} is able to reconstruct implicit surfaces from dense multi-view images while showing excellent performance in reconstructing fine-grained details.
Since then, several works have improved it in reconstruction quality \cite{yu2022monosdf}, sparse-view performance \cite{long2022sparseneus} and optimization speed \cite{wang2023neus2}.

Yet textureless and specular surfaces are still challenging for neural implicit surface reconstruction.
To address the problem, NeRO \cite{liu2023nero} incorporates inverse rendering to alleviate the shape-radiance ambiguity \cite{kaizhang2020} induced by specular reflection.
Ref-NeuS \cite{ge2023ref} borrows the idea of anomaly detection to decrease the weights of specular pixels during optimization and conditions the radiance on reflection direction in reflective scenes as in Ref-NeRF \cite{verbin2022ref}.
UniSDF \cite{wang2023unisdf} models diffuse and specular radiances in two MLPs separately.

Estimating a 3D radiance field and an SDF from 2D images is a highly ill-posed problem. During optimization, these methods may distort shapes to fit the reflection colors, causing severe reconstruction errors, or, as is commonly referred to, the shape-radiance ambiguity \cite{kaizhang2020}.
In contrast, polarization images provide linear constraints on surface normal by the angle of polarization.
Therefore, to avoid the shape-radiance ambiguity, our method utilizes the geometry cues from polarization to regularize object shapes independently of appearance, keeping the shape from distortions induced by the visually distorted appearance.

\subsubsection{Polarimetric 3D Reconstruction:}
Shape from polarization (SfP) \cite{rahmann2001reconstruction,miyazaki2003polarization, atkinson2006recovery, yu2017shape} and photo-polarimetric stereo \cite{ngo2015shape, tozza2021uncalibrated, ding2021polarimetric, ding2022polar} are single-view model-based methods that utilize polarization images for normal and depth estimation.
These methods rely on assumptions of either spotlighting or distant lighting to explicitly model the relationship between the degree of polarization and the zenith angle of a surface normal, making them difficult to work under natural lighting conditions.
Several deep learning-based methods \cite{ba2020deep, lyu2023pami, Shao_2023_ICCV} leverage data priors to enhance the robustness to more complex indoor lighting conditions.
SfPW \cite{Lei_2022_CVPR} and DPS-Net \cite{bao2023sine} learn to estimate scene-level normal maps and depth maps.
Yet their generalization is still limited by the relatively small amount of training datasets of polarization images.

Polarimetric multi-view 3D reconstruction leverages multi-view polarimetric constraints to adapt to natural lighting conditions.
The polarimetric constraint is established on the fundamental fact that the azimuth angle of a surface normal is equal to the angle of polarization up to a $\pi/2$-ambiguity and a $\pi$-ambiguity under natural lighting \cite{cui2017polarimetric, zhao2022polarimetric}, which is relatively independent of lighting \cite{cui2017polarimetric}.
Miyazaki \textit{et al.} \cite{miyazaki2012polarization} uses multi-view polarimetric constraints to reconstruct black specular objects.
PolMVS \cite{cui2017polarimetric} proposes to propagate the depths of sparse points and refine the depth maps with this constraint.
Inspired by PolMVS, PDMS \cite{yang2018polarimetric} and PMDM \cite{shakeri2021polarimetric} integrate this constraint into SLAM for dense mapping.
PMVIR \cite{zhao2022polarimetric} proposes a framework to simultaneously refine coarse meshes and estimate illuminations and surface albedos.
PolarPMS \cite{zhao2024polarimetric} combines this constraint with photo consistency to estimate depth maps from multi-view polarization images.

More closely related to our approach is PMVIR \cite{zhao2022polarimetric}.
PMVIR represents shapes using meshes and thereby has limitations in handling incorrect shape topology.
In contrast, by leveraging neural SDF, our method is able to rectify incorrect shape topology during the optimization.
More importantly, our method utilizes the perspective polarimetric constraint \cite{chen2022perspective} to consider the camera ray directions in the polarimetric loss and thus achieves further performance improvement.

\subsubsection{Polarimetric Inverse Rendering:}
The development of polarimetric bidirectional reflectance distribution function (pBRDF) \cite{baek2020image, kondo2020accurate, hwang2022sparse, ichikawa2023fresnel} has enabled polarimetric inverse rendering.
Based on mesh representation, sparse ellipsometry \cite{hwang2022sparse} leverages polarimetric inverse rendering to simultaneously estimate object meshes and materials using polarization images captured under flashlight illumination.

With the emergence of neural fields in 3D reconstruction, a few recent works \cite{dave2022pandora, li2023neisf} have integrated polarization into neural implicit surfaces through polarization image rendering.
Concurrent work NeISF \cite{li2023neisf} introduces the incident Stokes field to represent multi-path polarized light and reparameterizes pBRDF using MLPs, achieving highly accurate geometry reconstruction.
PANDORA \cite{dave2022pandora} aims to decompose diffuse and specular reflections using polarization as auxiliary cues, relying on the Fresnel equations to calculate the diffuse and specular components of radiance.

Although polarimetric inverse rendering excels at accurate material and illumination estimation, it requires proper geometry initialization to decouple shape and appearance.
As a result, it may still suffer from the shape-radiance ambiguity, especially when dealing with textureless and specular surfaces.
To avoid this problem, our method utilizes the polarimetric constraint to regularize shape directly, independent of appearance.
The surface reconstruction results of our method can be used to initialize polarimetric inverse rendering.

\section{Preliminary}
\subsubsection{Polarimetric Constraint:}
\label{sec:pac}
Assuming four synchronized and aligned images captured at polarizer angles of $0$, $45$, $90$, $135$ degrees \cite{polsensor},
denoted as $\mathbf{I_0}$, $\mathbf{I_{45}}$, $\mathbf{I_{90}}$ and $\mathbf{I_{135}}$.
For each pixel, the angle of polarization (AoP) $\varphi\in[-\pi/2,\pi/2]$ and degree of polarization (DoP) $\rho\in [0,1]$ are calculated as:
\begin{equation}\label{eq:aop}
\varphi=\frac{1}{2}\text{arctan2}(s_2,\ s_1),\quad
\rho = \frac{\sqrt{s_1^2+s_2^2}}{s_0}
\end{equation}
where $\mathbf{s}=[s_0,s_1,s_2]^T=[\frac{1}{2}\sum I_i,\;I_{0}-I_{90},\;I_{45}-I_{135}]^T$ is the Stokes vector for representing the polarization state of light.
Given the estimated normal $\hat{\mathbf{n}}$ in the camera frame and the AoP $\varphi$ of a pixel, the orthographic polarimetric constraint on the azimuth angle of the normal is calculated as:
\begin{equation}\label{eq:opa}
[\sin(\varphi+\Delta),\ \cos(\varphi+\Delta),\ 0]\cdot \hat{\mathbf{n}}=0,
\end{equation}
where the disambiguation scalar $\Delta\in\{0,\ \pi/2\}$ is used for solving the $\pi/2$-ambiguity\cite{cui2017polarimetric}.
While this constraint is commonly used in previous works \cite{cui2017polarimetric,yang2018polarimetric,shakeri2021polarimetric,tozza2021uncalibrated,zhao2022polarimetric}, there is a perspective polarimetric constraint which is geometrically more accurate than it by considering the perspective effect of the lens \cite{chen2022perspective}:
\begin{equation}\label{eq:ppa}
[v_z\sin \varphi',\ v_z\cos\varphi',\ 
-(v_y\cos\varphi'+v_x\sin\varphi')]\cdot \hat{\mathbf{n}}=0,\quad \varphi'=\varphi+\Delta
\end{equation}
where $\mathbf{v}=[v_x, v_y, v_z]^T$ is a normalized camera ray in the camera frame.

\subsubsection{$\pi/2$ ambiguity:}
\label{sec:ambi}
The $\pi/2$-ambiguity arises from the differences between the AoPs of specular reflection and diffuse reflection \cite{cui2017polarimetric}.
According to the Fresnel equations, reflected and transmitted light have distinct energy distributions in the parallel and perpendicular directions to the plane of incident, resulting in a $\pi/2$ difference between their AoPs, i.e., $\Delta=0$ for diffuse reflection and $\Delta=\pi/2$ for specular reflection.
Under natural lighting conditions, the diffuse reflections and specular reflections are usually mixtured, making it difficult to distinguish the dominant component, and thus resulting in the $\pi/2$-ambiguity.
Therefore, the $\pi/2$-ambiguity poses a challenge in terms of applying the polarimetric constraints.

\section{Method}
\label{sec:method}
\begin{figure}[t]
	\centering
	\includegraphics[width=122mm]{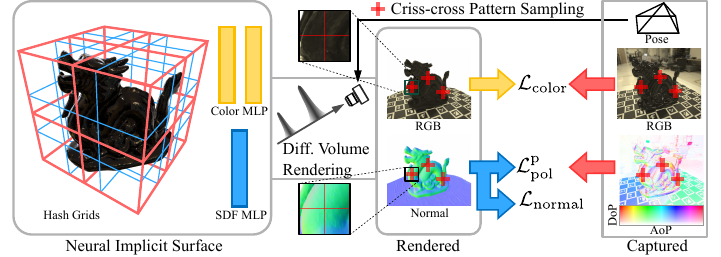}
	\caption{
		Overview of PISR.
		Pixels for the optimization are sampled in criss-cross patterns at the middle stage of the optimization.
		The polarimetric loss $\mathcal{L}^\text{p}_\text{pol}$ and the normal loss $\mathcal{L}_\text{normal}$ are used for regularizing the shape independently of appearance.
	}
	\label{fig:ch4-pipeline}
\end{figure}

PISR aims to recover object shapes from multi-view polarization images and camera poses.
\cref{fig:ch4-pipeline} shows an overview of our method.
Object shape is represented as a hash-grid-based neural SDF and
rendered to images for computing losses (\cref{sec:method-surfrend}).
During the reconstruction, the coarse shape of an object is estimated with the photometric loss $\mathcal{L}_{\text{color}}$.
Then the polarimetric loss $\mathcal{L}^\text{p}_\text{pol}$ (\cref{sec:method-apaloss}) and the normal loss $\mathcal{L}_\text{normal}$ with criss-cross pattern sampling (\cref{sec:method-smooth}) are jointly used to rectify the shape from distortions.
Finally, the shape is refined with $\mathcal{L}_{\text{color}}$ and $\mathcal{L}^\text{p}_\text{pol}$.
The optimization scheme is introduced in \cref{sec:method-scheme}.
\subsection{Surface Representation and Image Rendering}
\label{sec:method-surfrend}
Since Instant-NGP\cite{muller2022instant}, combining a multi-resolution hash grid and MLPs has been commonly used for faster convergence in neural field reconstruction.
Following recent works\cite{li2023neuralangelo,wang2023co,wang2023neus2}, we represent the shape and appearance of an object as an $L$-level multi-resolution hash grid $\mathcal{G}=\{\mathcal{G}_l\}^L_{l=1}$ with two MLPs $\Phi_s$ and $\Phi_\mathbf{c}$ for decoding SDF values $s$ and color $\mathbf{c}$ respectively.
For each 3D space point $\mathbf{x}$, the SDF value $s$ and color $\mathbf{c}$ are computed as
\begin{equation}\label{eq:sdf_color}
    (s, \mathbf{f})=\Phi_s(\mathcal{G}(\mathbf{x})),\quad
    \mathbf{c}=\Phi_\mathbf{c}(s, \mathbf{f}, \mathbf{v}, \mathbf{n})
\end{equation}
with the camera ray $\mathbf{v}$ and surface normal $\mathbf{n}$.
The normal $\mathbf{n}$ can be computed by normalizing the gradient of SDF w.r.t. to $\mathbf{x}$ or by finite difference\cite{li2023neuralangelo}.
Besides the foreground object, we use Instant-NGP \cite{muller2022instant} to represent the background.

To optimize the shape with images, we use differentiable volume rendering \cite{martin2021nerf} to render images given the surface representation.
The pixel color $\hat{\mathbf{c}}$ and normal $\hat{\mathbf{n}}$ are computed as the weighted sum of $M$ samples $\{\mathbf{x}_i=\mathbf{o}+t_i\mathbf{v}\}^M_{i=1}$ on a cast-off pixel ray
with the camera position $\mathbf{o}$ and different depth $t_i$:
\begin{equation}\label{eq:color}
    \hat{\mathbf{c}} = \sum_{i=1}^M w_i\mathbf{c}_i, \quad
    \hat{\mathbf{n}} = \sum_{i=1}^M w_i\mathbf{n}_i,
\end{equation}
where $w_i$ is the blending weight converted from SDF values using the unbiased weight function of NeuS \cite{wang2021neus}, and $\mathbf{c}_i$ and $\mathbf{n}_i$ are the color and normal of the sample $\mathbf{x}_i$.

\subsection{Polarimetric and Photometric Losses}
\label{sec:method-apaloss}

\noindent\textbf{Polarimetric Loss:}
Previous works \cite{cui2017polarimetric,yang2018polarimetric,shakeri2021polarimetric,tozza2021uncalibrated,zhao2022polarimetric} leverage polarization cues with the following loss function $h^\text{o}$ that is derived from the orthographic polarimetric constraint \cref{eq:opa}:
\begin{equation}\label{eq:hopa}
h^\text{o}(\varphi,\Delta,\mathbf{n})=
\Big(\mathbf{a}^\text{o}(\varphi,\Delta)\cdot \mathbf{n}\Big)^2,\;
\mathbf{a}^\text{o}(\varphi,\Delta)=[\sin(\varphi+\Delta),\ \cos(\varphi+\Delta),\ 0\ ],
\end{equation}
where the coefficient vector $\mathbf{a}^\text{o}$ is calculated with the AoP $\varphi$ and the disambiguation scalar $\Delta$.
However, as discussed in \cite{chen2022perspective}, \cref{eq:opa} neglects the perspective effect of the lens and may introduce distortions in the reconstruction results.

Therefore, we instead adopt the perspective polarimetric constraint \cref{eq:ppa} \cite{chen2022perspective} and propose a novel loss function $h^\text{p}$:
\begin{equation}\label{eq:hppa}
h^\text{p}(\varphi, \Delta, \mathbf{v}, \mathbf{n})=
\Big(\frac{\mathbf{a}^\text{p}(\varphi, \Delta,\mathbf{v})\cdot \mathbf{n}}{\|\mathbf{a}^\text{p}(\varphi, \Delta, \mathbf{v})\|}\Big)^2
\end{equation}
with $\mathbf{a}^\text{p}$ calculated as:
\begin{equation}
\mathbf{a}^\text{p}(\varphi, \Delta, \mathbf{v})=[v_z\sin\varphi',\ v_z\cos \varphi',\ 
-(v_y\cos \varphi'+v_x\sin \varphi')],\ 
\varphi'=\varphi+\Delta.
\end{equation}
The loss function $h^\text{p}$ additionally considers the camera ray direction $\mathbf{v}$ to model the perspective effect.
In \cref{sec:ch5-ablation}, we will show that $h^\text{p}$ effectively improves the reconstruction accuracy.

Besides, the $\pi/2$-ambiguity must be addressed when applying the polarimetric constraint.
We combine the merits of PolMVS \cite{cui2017polarimetric} and PMVIR \cite{zhao2022polarimetric}, proposing the segmented loss function $f^\text{p}$ as follows:
\begin{equation}\label{eq:fppa}
f^\text{p}(\varphi, \mathbf{v}, \mathbf{n}, \rho)=
\left\{
\begin{array}{ll}
h^\text{p}(\varphi, 0, \mathbf{v}, \mathbf{n})\cdot h^\text{p}(\varphi,\pi/2, \mathbf{v}, \mathbf{n}) & \text{ if } \rho<\theta,\\
\\
h^\text{p}(\varphi,\pi/2, \mathbf{v}, \mathbf{n}) &  \text{ if } \rho\ge\theta,
\end{array}
\right. 
\end{equation}
where $\rho$ is the DoP and $\theta$ is a preset threshold.
This design is based on two observations:
1) the reflected light of an object is usually a mixture of the diffuse and specular components, and thus the DoP is relatively low \cite{cui2017polarimetric},
2) only the DoP of specular reflection of dielectric objects can exceed $0.4$ according to Fresnel equations.
We use multiplication instead of addition when $\rho<\theta$ because only one of the two needs to be minimized.

Finally, we calculate the polarimetric loss for a batch of sampled pixels $\mathcal{S}$ as follows:
\begin{equation}\label{eq:apaloss}
    \mathcal{L}^\text{p}_{\text{pol}}=\frac{1}{|S|}\sum_{u\in \mathcal{S}} f^\text{p}(\varphi_u, \mathbf{v}_u, \hat{\mathbf{n}}_u, \rho_u).
\end{equation}
Note that the camera ray $\mathbf{v}_u$ and normal $\hat{\mathbf{n}}_u$ are transformed into the camera frame according to the camera pose.
In \cref{sec:ch5-ablation}, the experiments will demonstrate that calculating the loss $\mathcal{L}^\text{p}_{\text{pol}}$ with $h^\text{p}$ instead of $h^\text{o}$ can reduce the L1 Chamfer distance by $30$\%.

\smallskip\noindent\textbf{Photometric Loss:}
Since the polarimetric loss alone is insufficient for recovering object shapes due to the $\pi/2$ ambiguity and image noises,
we used the photometric loss for initializing and maintaining the basic shape:
\begin{equation}\label{eq:photoloss}
    \mathcal{L}_{\text{color}}=\frac{1}{|\mathcal{S}|}
    \sum_{u\in \mathcal{S}}|\hat{\mathbf{c}}_u-\mathbf{c}_u |.
\end{equation}

\subsection{Normal Regularization}
\label{sec:method-smooth}
The discrete nature of the hash grid makes the estimated SDF less smooth than those purely MLP-based, which can introduce reconstruction artifacts such as holes or cracks.
Directly smoothing the normals in 3D space as in \cite{zhang2021nerfactor,li2023neuralangelo}
can lead to a bumpy surface because the sampled normals are solely constrained by the smoothness loss.
Inspired by the belief propagation used in traditional MVS methods \cite{Zheng_2014_CVPR, schoenberger2016sfm, Xu_2019_CVPR},
we smooth the rendered normals so that all the samples are also constrained by both the photometric loss and the polarimetric loss.
We therefore utilize the following normal loss:
\begin{equation}\label{eq:smoothloss}
    \mathcal{L}_{\text{normal}}=
        -
        \frac{1}{|\mathcal{S}_\text{c}|}\sum_{u\in \mathcal{S}_\text{c}}
        \frac{1}{|\mathcal{N}_u|}
        \sum_{\hat{\mathbf{n}}_j\in\mathcal{N}_u}
        \text{SG}(\hat{\mathbf{n}}_u)^{T}\hat{\mathbf{n}}_j
\end{equation}
where $\mathcal{S}_\text{c}$ is a set of pixels sampled among pixels of all the images,
$\mathcal{N}_u$ is a set of pixel normals sampled around $\hat{\mathbf{n}}_u$ in the image space.
The function $\text{SG}$ (``Stop Gradient'') is used to stop propagating the gradients of the losses w.r.t. $\hat{\mathbf{n}}_u$ during backpropagation,
i.e., fix the values of $\hat{\mathbf{n}}_u$ so that it can be propagated to neighbor pixels.

Now the pixel sample set $S$ of each iteration during the optimization is the union of the set of center pixels $\mathcal{S}_c$ and the sets of neighbor pixels $\mathcal{N}_u$:
\begin{equation}\label{eq:sample}
    \mathcal{S}=\mathcal{S}_c\cup\{\bigcup_{u\in S_c}\mathcal{N}_u\}
\end{equation}
Empirically, larger $\mathcal{S}_\text{c}$ can lead to better results because it helps capture global structures, which are important for optimizing a neural field \cite{xie2023s3im},
while larger $\mathcal{N}_u$ is beneficial to smooth the surface.
Therefore, we sample $\mathcal{N}_u$ in a criss-cross pattern instead of a whole patch around $\hat{\mathbf{n}}_u$ to balance $|\mathcal{S}_\text{c}|$ and $|\mathcal{N}_u|$ with fixed $|S|$.
In principle, $\mathcal{N}_u$ can be sampled in other relatively sparse patterns.
Here we use the criss-cross one for its simple implementation in code.

\subsection{Optimization Strategy}
\label{sec:method-scheme}
The overall loss is
\begin{equation}\label{eq:all_loss}
    \mathcal{L}_{\text{all}}=
        \mathcal{L}_{\text{color}}
        +\lambda_\text{p}\mathcal{L}^\text{p}_{\text{pol}}
        +\lambda_\text{n}\mathcal{L}_{\text{normal}}
        +\lambda_\text{e}\mathcal{L}_{\text{eikonal}}
\end{equation}
The extra Eikonal loss $\mathcal{L}_{\text{eikonal}}=\frac{1}{M|S|}\sum^{M|S|}_{k=1}(\|\nabla\hat{s}(\mathbf{x}_k)\|-1)^2$ is used for encouraging the neural field to approximate a signed distance field \cite{gropp2020implicit}.
During the optimization, we apply progressive training \cite{li2023neuralangelo} on the multi-resolution hash grid.
To initialize a coarse shape, we set $\lambda_\text{p}=\lambda_\text{n}=0$ and use the photometric and eikonal losses at the early stage.
Then we gradually increase $\lambda_\text{p}$ and $\lambda_\text{n}$ to rectify the distorted shape.
Finally, we gradually decrease $\lambda_\text{n}$ to $0$ to avoid over smoothing the details.

\section{Experiments}
\label{sec:exps}
\subsection{Experimental Setup}
\begin{figure}[t]
    \centering
	\subfloat[\label{fig:a}]{\includegraphics[height=35mm]{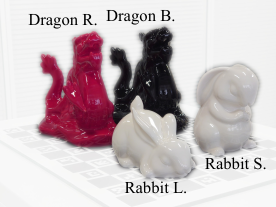}}\quad
	\subfloat[\label{fig:b}]{\includegraphics[height=35mm]{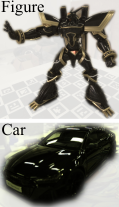}}\quad
	\subfloat[\label{fig:c}]{\includegraphics[height=35mm]{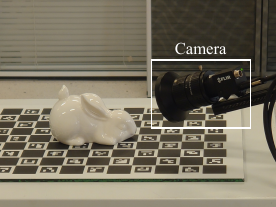}}
    \caption{Dataset and collection. (a) Objects with ground-truth meshes. (b) Objects without ground-truth meshes. (c) Capture setup.}
	\label{fig:capture}
\end{figure}

\noindent\textbf{Dataset:}
Since there is no public polarization image dataset of textureless and specular objects with ground-truth meshes, we collect a real-world dataset, including four 
 objects (\textit{Black Dragon}, \textit{Red Dragon}, \textit{Standing Rabbit}, \textit{Lying Rabbit}, \textit{Figure} and \textit{Car}) as shown in \cref{fig:capture}.
The reasons for capturing real object images, rather than synthesizing using a rendering engine, such as \cite{jakob2022mitsuba3}, are two-fold:
1) the natural ambient lighting conditions are complex and therefore difficult to synthesize,
and 2) the shadows and occlusions of the camera can be realistically generated.

Each object's polarization images are captured indoors under natural lighting conditions from 40 (60 for \textit{Figure} for its relatively complex shape) different viewpoints around the object with a color polarization camera\cite{polsensor}.
Color images are obtained by first decomposing polarization images into images of four polarizer angles and then averaging and demosaicing.
Camera poses are estimated from color images using SuperGlue \cite{sarlin2020superglue} and COLMAP \cite{schoenberger2016sfm} given pre-calibrated camera intrinsics.
We use a checkerboard to help pose estimation.
To obtain ground-truth meshes, the objects are coated by 3D scanning spray to eliminate specular reflection and scanned by a structure light-based scanner except for \textit{Figure} and \textit{Car}.
Please refer to our supplementary material for more details.\\

\noindent\textbf{Performance Metrics:}
Since we focus on geometry reconstruction, we only evaluate the surface reconstruction quality using L1 Chamfer distance and F-score.
We remove the background meshes before calculating the metrics.\\
\begin{figure}[t]
	\centering
	\subfloat{\includegraphics[width=122mm]{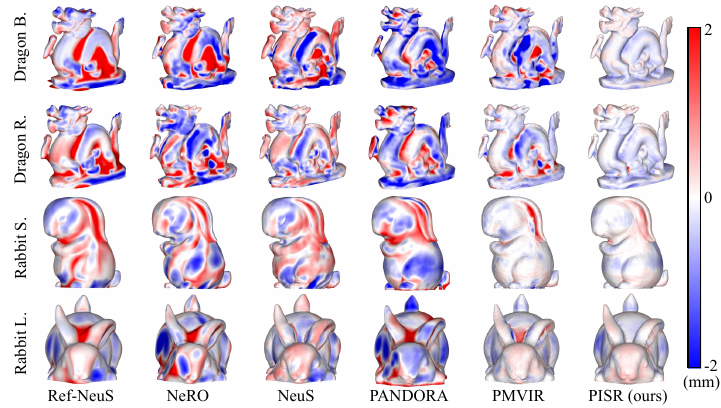}}
	\caption{
		Signed error maps.
		Red and blue mean surface swelling and shrinking respectively.
		Regions with higher color saturation indicate longer distances between the reconstruction result and the ground truth.
		Errors are truncated to within $\pm 2$mm for better visualization.
	}
	\label{fig:ch5-comp_qual}
\end{figure}
\begin{table}[t]
	\caption{
		Summary of Methods in Experiments
	}
	\label{table:baselines}
	\centering
	\renewcommand\tabcolsep{4pt}
	\renewcommand\arraystretch{1.0}
	\begin{tabular}{lccc}
		\toprule
		Methods & Image & Shape Repre. & Appearance Repre. \\
		\midrule
		Ref-NeuS \cite{ge2023ref} & RGB  & Neural SDF  &  Neural radiance      \\
		NeRO \cite{liu2023nero} &  RGB   & Neural SDF &    Neural BRDF    \\
		NeuS \cite{wang2021neus} &   RGB   & Neural SDF & Neural radiance     \\
		PANDORA \cite{dave2022pandora} & Pol. RGB  & Neural SDF  &   Neural BRDF     \\
		PMVIR \cite{zhao2022polarimetric} &  Pol. RGB   & Mesh & Spherical harmonics     \\
		PISR (Ours) &  Pol. RGB   &  Grids+Neural SDF  & Neural radiance   \\
		\bottomrule
	\end{tabular}
\end{table}

\noindent\textbf{Implementation Details:}
The SDF MLP $\Phi_s$ has one hidden layer with a size of 64, and the color MLP $\Phi_\mathbf{c}$ has two hidden layers with sizes of 64.
The multi-resolution grid contains 16 levels of resolution, and the coarsest and finest resolutions are $32$ and $2700$, respectively.
The hash table size of each level is $2^{19}$.
The optimization of the grid starts at the fourth-coarsest resolutions and progressively extends one level every $1000$ iterations.
Throughout the whole optimization process, 
the threshold of DoP $\theta$ is set to 0.3 and
$\lambda_{\text{eikonal}}$ is set to $0.1$. 
After the first $2.5$k iterations, $\lambda_\text{pol}$, $\lambda_\text{normal}$ and $|\mathcal{N}_p|$ are linearly increased from $0$ to $2.0$, $1.0$ and $28$ in $2.5$k iterations, respectively.
Then $\lambda_\text{normal}$ and $|\mathcal{N}_p|$ are linearly decreased to $0$ in another $2.5$k iterations.
The maximum number of iterations is $20$k.
The final SDF is converted to meshes using Marching Cubes\cite{lorensen1987marching}.\\

\noindent\textbf{Baselines:}
We compare our method with five methods, including PMVIR \cite{zhao2022polarimetric}, PANDORA \cite{dave2022pandora}, NeRO\cite{liu2023nero}, Ref-NeuS \cite{ge2023ref} and NeuS \cite{wang2021neus}.
\cref{table:baselines} summarizes the type of input images, geometry representation and color representation of all the six methods.
Since PMVIR is a refinement method based on an initial shape, we provide the coarse initial meshes reconstructed by PISR-C (see the ablation study in \cref{sec:ch5-ablation} for the details of PISR-C).

\subsection{Results}
\noindent\textbf{Quantitative Results:}
The Chamfer distances and F-scores of the reconstruction results are reported in \cref{table:quan_comp}.
Our method achieves the best reconstruction quality on all four objects shown in \cref{fig:a} and significantly outperforms the other five methods.
The average Chamfer distance and F-score of our method are $0.5$ mm and $99.5\%$, respectively.
As the second-best method in the comparison, PMVIR achieves twice the Chamfer distance and a $9\%$ lower F-score compared to ours.
Both PMVIR and our method utilize polarimetric constraints, and the results show that the geometry cue of polarization is effective for shape estimation.
Although PANDORA also uses polarization images as input, it implicitly utilizes polarization cues through polarimetric rendering, i.e., optimizing shapes using an image reconstruction loss, which could suffer from the shape-radiance ambiguity as RGB-based methods.
Among RGB-based methods, NeuS achieves better results than Ref-NeuS and NeRO which are designed for specular objects.
\begin{table}[t]
\caption{
    Reconstruction accuracy in Chamfer distance (CD$\downarrow$, mm) and F-score (FS$\uparrow$, \%).
    FS is calculated at a threshold of $1$ mm.
    Note that the values in the table are rounded to one decimal place.
    The CD of PMVIR and PISR on Rabbit L. are $0.584$ mm and $0.578$ mm respectively.
    \textbf{Bold}, \uline{underline}, and \uuline{double underline} mean the best, second-best and third-best performance, respectively.
}
\label{table:quan_comp}
\centering
\renewcommand\tabcolsep{4pt}
\renewcommand\arraystretch{1.0}
\begin{tabular}{c|lcccccccccc}
\toprule
\multicolumn{2}{c}{\multirow{2}{*}{Methods}} & \multicolumn{2}{c}{Dragon B.} & \multicolumn{2}{c}{Dragon R.} & \multicolumn{2}{c}{Rabbit S.} 
 & \multicolumn{2}{c}{Rabbit L.} & \multicolumn{2}{c}{Avg.}  \\
 \cmidrule(lr){3-4} \cmidrule(lr){5-6} \cmidrule(lr){7-8} \cmidrule(lr){9-10} \cmidrule(lr){11-12} 
\multicolumn{2}{l}{} & CD & FS & CD & FS & CD & FS & CD & FS & CD & FS \\

 \midrule
\multirow{3}{*}{\rotatebox{90}{RGB}} & Ref-NeuS & 2.1 & 65.3 & 2.1 & 66.1 & 1.2 & 84.5 & 1.0 & 90.5 & 1.6 & 76.6 \\
 & NeRO & 2.4 & 59.2 & 1.8 & 66.3 & 1.3 & 80.5 & 1.3 & 85.0 & 1.7 & 72.8 \\
 & NeuS & \uuline{1.9} & \uuline{67.1} & \uuline{1.2} & \uuline{86.0} & \uuline{1.1} & \uuline{86.8} & \uuline{0.9} & \uuline{93.0} & \uuline{1.3} & \uuline{83.2} \\
\hline
\multirow{3}{*}{\rotatebox{90}{P. RGB}} & PANDORA & 2.3 & 55.3 & 1.8 & 64.4 & 1.4 & 78.3 & 1.5 & 77.1 & 1.8 & 68.8 \\
 & PMVIR & \uline{1.8} & \uline{76.5} & \uline{1.0} & \uline{90.5} & \uline{0.6} & \uline{97.5} & \uline{0.6} & \uline{97.3} & \uline{1.0} & \uline{90.5} \\
 & PISR (Ours) & \textbf{0.5} & \textbf{99.9} & \textbf{0.6} & \textbf{99.1} & \textbf{0.5} & \textbf{99.7} & \textbf{0.6} & \textbf{99.4} & \textbf{0.5} & \textbf{99.5} \\

\bottomrule
\end{tabular}
\end{table}

\begin{figure}[t]
	\centering
	\subfloat{\includegraphics[width=122mm]{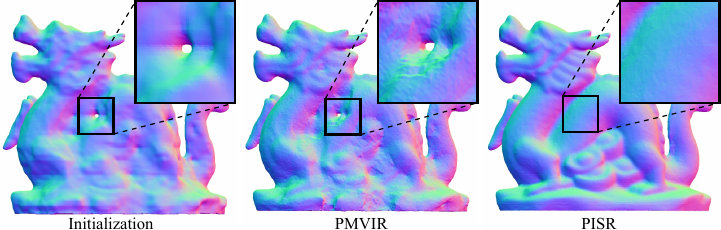}}
	\caption{
		Normal maps of Dragon B. before and after optimizations.
		Using neural SDFs to represent shapes allows changes in the shape topology during the optimization.
	}
	\label{fig:ch5-holes}
\end{figure}
\begin{table}[ht]
	\caption
	{
		Average optimization time of different methods.
	}
	\label{table:speed}
	\centering
	\renewcommand\tabcolsep{4pt}
	\renewcommand\arraystretch{1.0}
	\begin{tabular}{lcccccc}
		\toprule
		Methods & Ref-NeuS & NeRO & NeuS & PANDORA & PMVIR & PISR (Ours) \\
		\midrule
		Iterations & 200k & 200k & 200k & 100k & 4 & 20k \\
		\hline
		Time (h) &   20.0   &    20.5    &   10.8   &   15.0  &     2.0     &   0.5   \\
		\bottomrule
	\end{tabular}
\end{table}

We visualize the error distributions in \cref{fig:ch5-comp_qual}.
For methods that rely solely on image reconstruction losses (Ref-NeuS, NeRO, NeuS and PANDORA), their reconstruction errors in non-convex regions are notably higher compared to PMVIR and our method.
While the inherent continuity of MLPs ensures the watertightness of their reconstructed shapes, their accuracy cannot be guaranteed due to the shape-radiance ambiguity introduced by specular reflections and the lack of textures.
Additionally, for NeRO and PANDORA, the lighting shadows cast by the camera further exacerbate the difficulty of estimating surface material parameters.
The real HDR ambient lighting also poses challenges for PANDORA and Ref-NeuS to accurately estimate object shapes.
In contrast, PMVIR and our method can recover non-convex regions using the polarimetric constraints.
Despite the polarization cues, representing the shapes as neural SDF helps our method better reconstruct regions that are poorly initialized.
As shown in \cref{fig:ch5-holes}, neural SDF allows continuous changes of the shape topology during optimization, which is crucial for evolving surfaces from holes.\\

\noindent\textbf{Optimization Time:}
We report the average optimization time in \cref{table:speed}.
The experiments are conducted on a machine equipped with an Intel Core i9-12900K and an NVIDIA RTX 3090 GPU.
We set the maximum number of iterations to 200k.
If the shape converges, we stop the optimization earlier.
Our method represents shape as a multi-resolution hash grid and two shallow MLPs, significantly reducing the time complexity for rendering a pixel.
Therefore, our method converges much faster, achieving up to $4\sim41\times$ convergence speed while maintaining the highest reconstruction accuracy.

\subsection{Ablation Study}
\begin{table}[t]
\caption{
    Results of the ablation study on different combinations of losses.
    The L1 CD$\downarrow$ (mm) and FSs$\uparrow$ (\%) at $1.0$ mm and $0.5$ are the averages over four objects in Fig.\ref{fig:a}.
    The ranks are the averages over the three metrics.
    $\mathcal{L}^\text{o}_\text{pol}$ differs from $\mathcal{L}^\text{p}_\text{pol}$ by its use of the orthographic constraint Eq. \ref{eq:opa} instead of the perspective one Eq. \ref{eq:ppa}.
    Note that $\mathcal{L}_{\text{eikonal}}$ is used in all experiments and is omitted it in this table.
}
\label{table:ablation}
\centering
\renewcommand\tabcolsep{4pt}
\renewcommand\arraystretch{1.0}
\begin{tabular}{lccccccccc}
\toprule
\multirow{2}{*}{Variants}  & \multicolumn{4}{c}{Used Losses} & \multicolumn{3}{c}{Avg.} & \multirow{2}{*}{Rank}\\
\cmidrule(lr){2-5} \cmidrule(lr){6-8}
& $\mathcal{L}_{\text{color}}$      &
$\mathcal{L}^\text{o}_\text{pol}$      &
$\mathcal{L}^\text{p}_\text{pol}$ &
$\mathcal{L}_{\text{normal}}$   & CD & FS(1.0) & FS(0.5) \\
\midrule
PISR-C & $\checkmark$ &              &              &              & 1.10  & 88.10 & 70.30  & 6 \\
PISR-N & $\checkmark$ &              &              & $\checkmark$ &  1.00   & 88.20 & 71.10  & 5 \\
\hline
PISR-O & $\checkmark$ & $\checkmark$ &              &              & 0.80 & \uuline{96.60} & \uuline{75.54} & 3\\
PISR-ON & $\checkmark$ & $\checkmark$ &              & $\checkmark$ & \uuline{0.77} & 96.10 & 75.50 & 4 \\
\hline
PISR-P & $\checkmark$ &              & $\checkmark$ &              & \textbf{0.51} & \textbf{99.76} & \textbf{92.79}  & 1\\
PISR(-PN) & $\checkmark$ &              & $\checkmark$ & $\checkmark$ & \uline{0.54}  & \uline{99.54} & \uline{91.75} & 2 \\
\bottomrule
\end{tabular}
\end{table}

\label{sec:ch5-ablation}
We evaluate the effectiveness of the polarimetric loss and normal regularization,
as well as the influence of the perspective effect of the lens.
We compare PISR(-PN) with its five variants, referred to as \textit{PISR-P}, \textit{PISR-ON}, \textit{PISR-O}, \textit{PISR-N} and \textit{PISR-C}, based on different combinations of losses as shown in \cref{table:ablation}.
The quantitative experiment is conducted on objects shown in \cref{fig:a} and the qualitative experiment is conducted on objects shown in \cref{fig:b}.

\begin{figure}[tb]
	\centering
	\subfloat{\includegraphics[width=122mm]{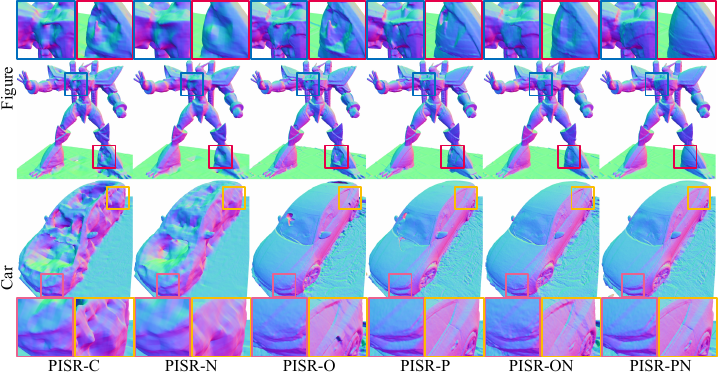}}
	\caption{
		Normal maps of Figure and Car.
		The polarimetric loss and normal loss jointly recover the shape from distortions induced by strong reflections.
	}
	\label{fig:ch5-ablation_qual}
\end{figure}

As shown in \cref{table:ablation}, the top four performing variants are all incorporated with the polarimetric losses, and fine-grained details can be recovered without complicated appearance modeling.
As for the normal regularization, although the quantitative comparison between PISR-PN and PISR-P shows that it slightly reduces the accuracy,
the qualitative results in \cref{fig:ch5-ablation_qual} show its effectiveness for recovering surfaces from severe distortions.
\begin{figure}[tb]
	\centering
	\subfloat{\includegraphics[height=35mm]{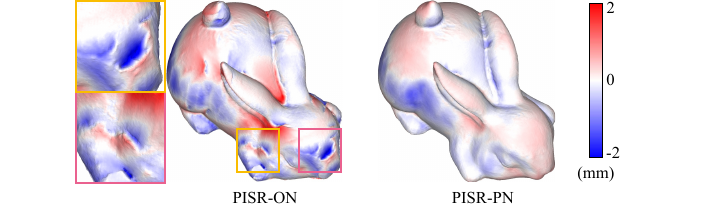}}
	\caption{
		Signed error maps of Rabbit L.
		PISR-PN leverages the perspective polarimetric constraint to eliminate surface cracks encountered by PISR-ON.
		Additional error maps can be found in our supplementary material.
	}
	\label{fig:ch5-ablation_ppa}
\end{figure}

Among the top four performing variants, PISR-PN and PISR-P achieve higher accuracy by considering the perspective effect of the lens using \cref{eq:ppa} instead of \cref{eq:opa}.
As the example error maps of Rabbit L. shown in \cref{fig:ch5-ablation_ppa}, the perspective effect leads to surface swelling and shrinking, causing severe cracks on the surface.
This phenomenon does not appear in the results of PMVIR and PANDORA.
We speculate the reason to be that the hash-grid-based neural SDF we use is too flexible in topology compared to the mesh-based one used in PMVIR, and lacks the smoothness of purely MLP-based one used in PANDORA.
Besides, the results in \cref{table:ablation} show that PISR-PN outperforms PISR-ON by a $30$\% reduction in L1 Chamfer distance and a $17\%$ improvement in F-score at $0.5$ mm,
which suggests the necessity of considering the perspective effect to improve the accuracy of polarimetric 3D reconstruction.

\section{Conclusion}
We have proposed PISR, an accurate and efficient polarimetric neural implicit surface reconstruction method for textureless and specular objects.
The key idea of PISR is to explicitly regularize the SDF using the polarimetric constraints to avoid the shape-radiance ambiguity.
More importantly, PISR achieves high accuracy by leveraging the perspective polarimetric constraint, eliminating the surface cracks caused by the traditional orthographic one.
Besides, PISR regularizes surface normals in image space for robustness and integrates the multi-resolution hash grid for efficiency.

\noindent\textbf{Limitations:}
Despite the high reconstruction accuracy in the experiments, PISR's performance degenerates with noisy AoP maps,
limiting its application in reconstructing rough, well-textured or mirror-like objects.
Recent works \cite{liu2023nero, li2023neisf} have shown solutions to mitigate this problem.
In addition, PISR also requires dense input views, adding shape constraints derived from normal and depth estimation models would be a promising solution.

%
%
\bibliographystyle{splncs04}
\bibliography{main}

\end{document}